\documentclass[10pt,twocolumn,letterpaper]{article}

\usepackage{iccv}
\usepackage{times}
\usepackage{epsfig}
\usepackage{graphicx}
\usepackage{amsmath}
\usepackage{amssymb}

\usepackage{multirow}
\usepackage{array}

\usepackage[pagebackref=true,breaklinks=true,letterpaper=true,colorlinks,bookmarks=false]{hyperref}

 \iccvfinalcopy 

\def\iccvPaperID{5} 

\ificcvfinal\fi
\begin{document}

\title{ Learning a 3D descriptor for cross-source point cloud registration from synthetic data}

\author{Xiaoshui Huang\\
University of Technology Sydney\\
{\tt\small Xiaoshui.Huang@student.uts.edu.au}
}

\maketitle

\begin{abstract}
As the development of 3D sensors, registration of 3D data (e.g. point cloud) coming from different kind of sensor is dispensable and shows great demanding. However, point cloud registration between different sensors is challenging because of the variant of density, missing data, different viewpoint, noise and outliers, and geometric transformation. In this paper, we propose a method to learn a 3D descriptor for finding the correspondent relations between these challenging point clouds. To train the deep learning framework, we use synthetic 3D point cloud as input. Starting from synthetic dataset, we use region-based sampling method to select reasonable, large and diverse training samples from synthetic samples. Then, we use data augmentation to extend our network be robust to rotation transformation. We focus our work on more general cases that point clouds coming from different sensors, named cross-source point cloud. The experiments show that our descriptor is not only able to generalize to new scenes, but also generalize to different sensors. The results demonstrate that the proposed method successfully aligns two 3D cross-source point clouds which outperforms state-of-the-art method. 
\end{abstract}

\section{Introduction}

Point cloud is an important type of geometric data structure, which is not affected by occlusion, illumination and geometric transformation. Because of this property, point cloud is becoming prevalent as a format to store 3D data. There recently has endured high development in 3D sensors. Nowadays, there are many sensors available to produce point clouds, such as KinectFusion, LiDAR, range camera. Cross-source point clouds are data coming from different kinds of sensors. As different sensor shows different ability in sensing our environment, for example, KinectFusion shows strong ability in small-scale scene with many details and LiDAR shows strong ability in large-scale scene, registration of these point clouds shows indispensable. Therefore, cross-source point cloud registration problem emerges. It shows many applications including automatic driving, smart city and smart phone application.

\begin{figure}
	\includegraphics[width=8cm,height=4cm]{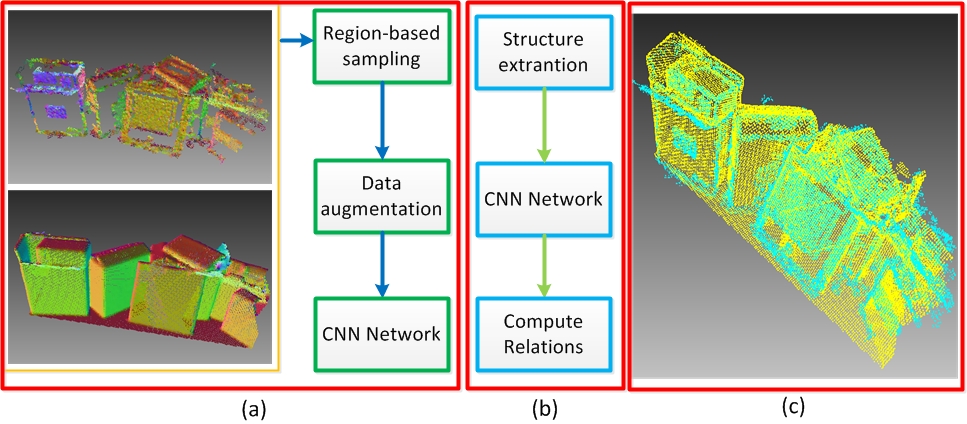}
	\caption{The outline of the proposed method. (a) shows the networking training method, with two cross-source point clouds inputing, the network is trained by using two strategies proposed in this paper, (b) shows one application on 3D cross-source point cloud registration, (c) shows the registration result. }
	\label{f1}
\end{figure}

The problem existing in cross-source point cloud registration are mainly lies in the variations on density, missing data, different viewpoints, scale, noise and outliers \cite{CSGM}. There is some papers in dealing with cross-source point cloud registration \cite{CSGM, cs-coarse-to-fine,7987811}. One of the key procedure of the exiting methods is the similarity measurement of two components. The exiting methods are mainly relied on handcraft features, such as graph descriptors \cite{CSGM}. Recently, learning methods (e.g. convolution neural network (CNN)) show great ability in dealing with different kinds of computer vision problems. In this paper, we propose a method to utilize the ability of learning methods to deal with 3D cross-source point clouds registration.

The main difficulty is that point cloud is usually organized as irregular format where each point represents a 3d position. This type of data is difficult to be directly utilized as input for neural network because of its irregular property ( examples of regular data are RGB-D data and images). Most researchers typically transform such data to regular 3D voxel grids or collections of images before feeding them to neural network architecture (e.g. \cite{3dshape, 3DMatch}). However, the limitations are summarized as two folders: 
\begin{enumerate}
	\item{they will render the resulting data unnecessarily voluminous while also introducing quantization artifacts that can obscure natural invariance of the data;}
	\item{the existing methods are sensitive to rotation variance.}
\end{enumerate}
The difficulties lying in learning on cross-source point cloud are that there is no stable keypoints in cross-source point cloud because of the many cross-source variations. It is difficult to find positive samples to train a neural network. 

Although there are many variations on cross-source point cloud, based on our observation, the structure of two point clouds remains similar. Inspired by this observation, in this paper, the function of structure similarity measurement is learned. If these two patches are describing a same structure, the similarity is very high. Otherwise, the similarity is very low. In order to fulfill our goal, we propose a region-based method to train a neural network.

In this paper, we propose a region-based sampling method to obtain training data for neural network architecture (learning framework) and utilize the original point clouds as input for neural networks. We also use data augmentation method to let our network be robust for point cloud rotation. Furthermore, Our work focuses on more general case where point cloud are coming from different sensors (cross-source point cloud). We try to learn 3D descriptors for cross-source point cloud to address cross-source 3D registration (See Figure \ref{f1}). 

The contributions are two folds: 1) a region-based method is proposed to deal with 3D point cloud training problem. It is robust to noise and outliers. 2) 3D rotation variant is dealt with by data augmentation.

\section{Related Works}

Learning a 3D descriptor for cross-source point cloud registration is a intersection area of computer vision and machine learning. We briefly review the related works on both domains.

\subsection{Handcraft 3D descriptors for point cloud}
Many 3D descriptors have been proposed including Fast Point Feature Histogram (FPFH) \cite{fpfh},  Signature of Histograms of Orientations (SHOT) \cite{shot}, and Clustered Viewpoint Feature Histogram (CVFH) \cite{cvfh}, Ensemble of Shape Functions (ESF) \cite{esf}. X. Huang \cite{CSGM} proposes relative geometry descriptor to describe graphs of 3D point cloud. Many of these 3D descriptors are now available in the Point Cloud Library \cite{pcl}. Although these existing 3D descriptors have make a great progress for 3D computer vision, they still  struggle to handle density variation, missing data, viewpoint difference variation, large proportion of noise and outliers in real-world data from  different 3D sensors. Furthermore, since different sensors usually have different sensing mechanisms and they manually designed for specific applications, the captured 3D data are usually have large variance due to their different sensors and different captured environment. In this paper, we propose a method to learn a 3D descriptor that directly learns from irregular 3D point cloud. The new learned 3D descriptor shows more robust and high discriminative in a variety of 3D applications.
 
\subsection{Learned 3D descriptors}
There has also been rapid progress in learning geometric representations on 3D  data. 3D ShapeNets \cite{3dshape} introduced 3D deep learning for  modeling 3D shapes, and several recent works \cite{song2016deep, maturana20153d, fang20153d}  also compute deep features from 3D data for the task of object retrieval and classification. While these works are inspiring, their focus is centered on extracting features from  complete 3D object models at a global level. In contrast,  our descriptor focuses on learning geometric features for irregular 3D point cloud at a local level, to provide  more robustness when dealing with partial data suffering from various occlusion patterns and viewpoint differences.

Guo \cite{guo20153d}, which uses a 2D ConvNet descriptor to match local geometric features for mesh labeling. \cite{3DMatch} learned from RGB-D data and utilize to deal with 3D point cloud feature extraction. However, these methods are all trained or designed for same sensors, they face large challenging when confront with different sensors' data with different data modalities. In contrast, our work not only  tackles the harder problem of matching real-world partial 3D point cloud, but also generalize our methods to different sensors in a spatially coherent way. Our method has higher robustness to different sensors' variants and shows ability to deal with rotation spatial transformation.

\section{Region-based method and data augmentation}
There are two main works in the proposed methods: region-based method for sample selection and data augmentation for addressing geometric rotation between cross-source point clouds. 

\begin{figure*}[ht]
	\centering
	\includegraphics[width=17cm,height=2cm]{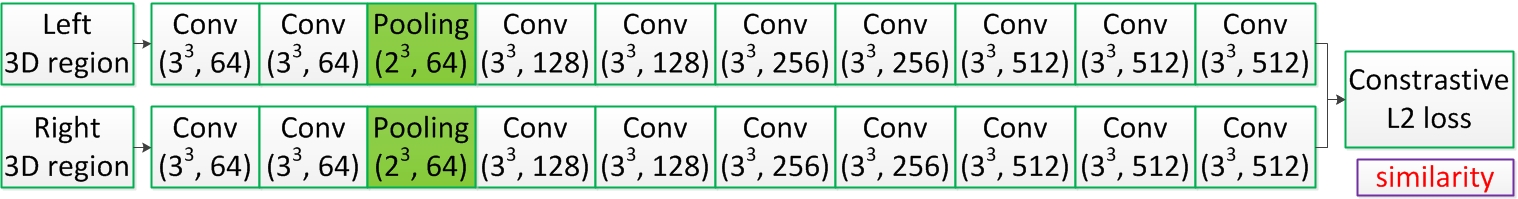}
	\caption{Network structure of our constitutional neural network. \textit{Conv} represents convolution layer, \textit{Pooling} represent pooling layer, the number behind each layer are the its parameters. $(3^3, 64)$ represents the kernel size is $3\times3\times3$ and the output size is 64.}
	\label{f3}
\end{figure*}
\begin{figure}
	\includegraphics[width=8cm,height=6cm]{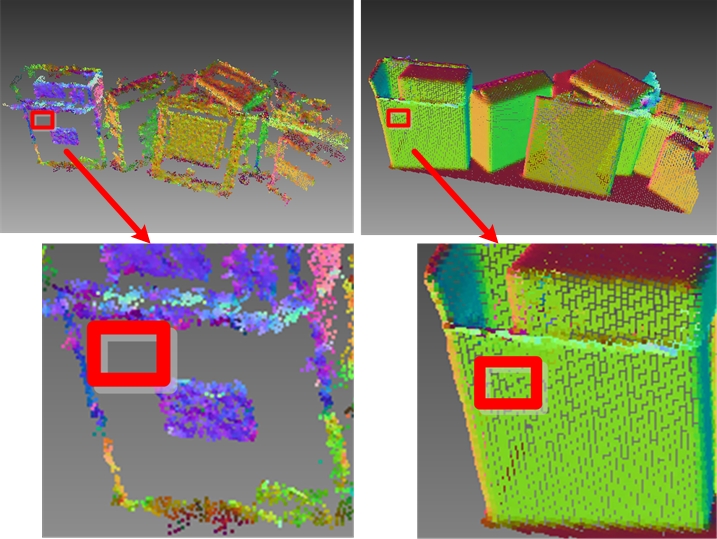}
	\caption{Example of deleted sample region. The red box are correspondent regions that are selected. Because left patch contains no points, this sample are not used for training.}
	\label{f2}
\end{figure}

\textit{Region-based method:} Supposing two cross-source point clouds PC1 and PC2, we try to use region-based method to construct positive and negative samples. Firstly, features (e.g. curvature, FPFH, normal or 3D coordinate) are computed for every point in PC1 and PC2. Secondly, starting from PC1, we randomly select a point and use the features of the point to cluster similar neighbor points based on different random threshold. If the number of clustering points is more than 50, the 3D box containing these points are recorded. We aim to keep each region to contain similar shape or structure. Thirdly, using this 3D box, we crop a region from PC2. If the number of points are more than 50, we consider these two regions from PC1 and PC2 as positive samples (matched regions). We also capture a negative samples (non-matched regions) which the center of its containing box is a distance away to this positive sample. The distance ranges from 3*r to max radius of point cloud. r is the radius of the 3D containing box. For each scene, we sample 100 thousand positive and 100 thousand negative samples to construct the training dataset.

\textit{Data augmentation:} In order to be robust to rotation, we use data augmentation to expend the original database. We randomly rotate an angle from $-90^\circ$ to $90^\circ$ for each positive and negative samples. After data augmentation, one positive sample becomes 2 positive samples and the same increment on negative samples.

The advantages of the proposed method are three folds: \textbf{First}, region-based sampling methods delete many regions that have very sparse points. These high sparse regions often are very hard to maintain the structure similarity (see red box of Figure \ref{f1}). Using the refined regions, the dataset are more robust to utilize the structure information.   \textbf{Second}, because the region-based method are usually has different region size, combined with high variance of cross-source point clouds, this method provides a large and diverse correspondence training data set. Training by our method, although the keypoints are difficult to extract, the network can still train a robust descriptor and be applied to real-world 3D cross-source point cloud registration (See our experimental part).  \textbf{Third}, due to the data augmentation, our network can be endowed with rotation ability. Therefore, the trained framework using the data augmentation samples can be robust to address the rotation of point clouds which is a common problem in many 3D applications.

\section{Network architecture}
The architecture of this paper is depicted in Figure \ref{f3}. The network (ConvNet) consists 10 layers with 9 convolution layers and 1 pooling layer. For each convolution layer, it is a standard 3D convolution layer, the only difference is the output size. For example, the convolution layer with parameter $(3^3, 64)$ represents this layer consists 64 kernels of size $3\times3\times3$. For pooling layer, it has 64 kernels with size $2\times2\times2$. Because we use the 3D region with variant size as inputs, the network can be trained with even larger and diverse dataset which has higher generalization.
	
\section{Training}
During training, our objective is to optimize the local descriptors generated by the proposed network (ConvNet) such that they are similar for 3D patches corresponding to the same point, and
dissimilar otherwise. To this end, we train our ConvNet with two streams in a Siamese fashion where each stream independently computes a descriptor for a different local 3D patch. The first stream takes in the local 3D patch around a surface point p1 , while the second stream takes in a second local 3D patch around a surface point p2 . Both streams share the same architecture and underlying weights. We use the L2 norm as a similarity metric between descriptors, modeled during training with the contrastive loss function \cite{chopra2005learning}. There is a margin super parameter in this loss. The margin is used to control the minimization  on positive 3D point pairs, while the negative 3D point pairs only work when the distance below the margin. In training process, the network has an input of a same number of positive and negative samples. With the same positive and negative samples feeding for the network, it has shown high efficiency to learn discriminative descriptors \cite{simo2015discriminative,yi2016lift,3DMatch}.

\begin{figure}[ht]
	\centering
	\includegraphics[width=8cm,height=1cm]{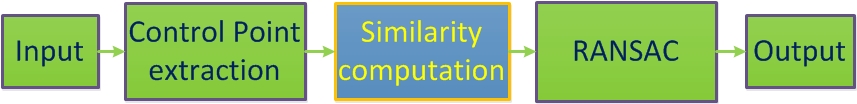}
	\caption{The pipeline of the cross-source point cloud registration. Our main contribution lying in similarity computation. }
	\label{f4}
\end{figure}

\begin{table*}[ht]
	\begin{center}
		\begin{tabular}{|l|c|c|c|c|c|c|c|c|}
			\hline
			Experiments & \multicolumn{2}{c|}{Position} &  \multicolumn{2}{c|}{CS problem} & \multicolumn{2}{c|}{Rotation} &  \multicolumn{2}{c|}{Result} \\
			\hline
			Methods & Ours & 3DMatch & Ours & 3DMatch& Ours & 3DMatch& Ours & 3DMatch\\
			\hline\hline
			Experiment 1 & ${\times}$ & ${\times}$   & ${\times}$&${\times}$   & ${\times}$&${\times}$   & 100\%&100\% \\
			Experiment 2 & ${\surd}$&${\surd}$ & ${\times}$&${\times}$   & ${\times}$&${\times}$   & 100\%&100\% \\
			Experiment 3 & ${\times}$&${\times}$   & ${\surd}$&${\surd}$ & ${\times}$&${\times}$   & 98.8\%&4.2\% \\
			Experiment 4 & ${\times}$&${\times}$   & ${\times}$&${\times}$   & ${\surd}$&${\surd}$ & 6\%&0\% \\
			\hline
		\end{tabular}
	\end{center}
	\caption{Experimental results on different components.}
	\label{t1}
\end{table*}
\section{Application}
We take one application as 3D cross-source point cloud registration. Figure \ref{f4} shows the pipeline of 3D cross-source point cloud registration. The inputs are two cross-source point clouds that needs to be registered. The outputs are the registration result and transformation matrix. In our pipeline, there are three main steps. Firstly, we extract the control points that robustly represent the structure of the point clouds. In this paper, we use 3D segmentation method \cite{3dvoxel} to segment the point cloud into many parts, the center of each segment is the control point. 
Secondly, we use the network in Figure \ref{f3} to extract the features and compare the similarity of these control points. In this step, we use the containing sphere of its belonging 3D segment as input and use Truncated Distance Function (TDF) in \cite{3DMatch} to convert the original point cloud into another 3D representation that suitable to do 3D convolution. Thirdly, we use simple 3D RANSAC to do the outliers detection and obtain the final transformation matrix.

\section{Experiments}
In this section, we conduct experiments and compare with the related work \cite{3DMatch}. We do three experiments, (1) one experiment is to compare the ability in dealing with cross-source problem; (2) the other experiment is to evaluate how well the proposed method in dealing with rotation and translation combined with cross-source problem; (3) apply to the real cross-source 3D point cloud registration.

\subsection{Compare the ability on cross-source problem}
To test the performance of the proposed method in dealing with cross-source problem, we separately evaluate and compare the ability in each variant and compare with \cite{3DMatch}. We test how our method is influenced by factors of position, cross-source problem and rotation. We have done four experiments: \textit{Experiment 1} is a baseline to test the accuracy of the algorithm. \textit{Experiment 2} aims to test different position impacts the descriptor. \textit{Experiment 3} aims to test cross-source problem impacts the descriptor. \textit{Experiment 4} aims to test rotation impacts the descriptor.

We use overlap ratio to evaluate the final registration accuracy. For overlap ratio, we compute nearest neighbor for each point, then count the number of point whose distance less than 0.05. We use the number divide by the total point number as the overlap ratio.
So, $overlapratio=N/N_t$. $N$ is the number whose nearest distance less than 0.05, $N_t$ is the total point number.

Table \ref{t1} shows that both methods obtain 100\% overlap ratio in Experiment 1 and Experiment 2. It reflect both our methods can be used to deal with 3D point cloud and invariant to translation variant. Experiment 3 shows our method is much more robust than \cite{3DMatch} when facing cross-source problem, the overlapping ratio improves 94\%. Experiment 4 shows our method still remain 6\% overlapping ratior while \cite{3DMatch} totally fails. Compared with \cite{3DMatch}, this experiment shows the proposed method shows higher ability in cross-source problem and can deal with part of the rotation problem.

\subsection{Evaluate and compare the ability on transformation variant}
In this section, we will conduct thorough experiments on transformation variants, which include experiments on rotation, shifting, rotation+shifting. 
Now, I will describe details about how do I conduct these experiments.

\begin{figure}[ht]
	\centering
	\includegraphics[width=8cm,height=6cm]{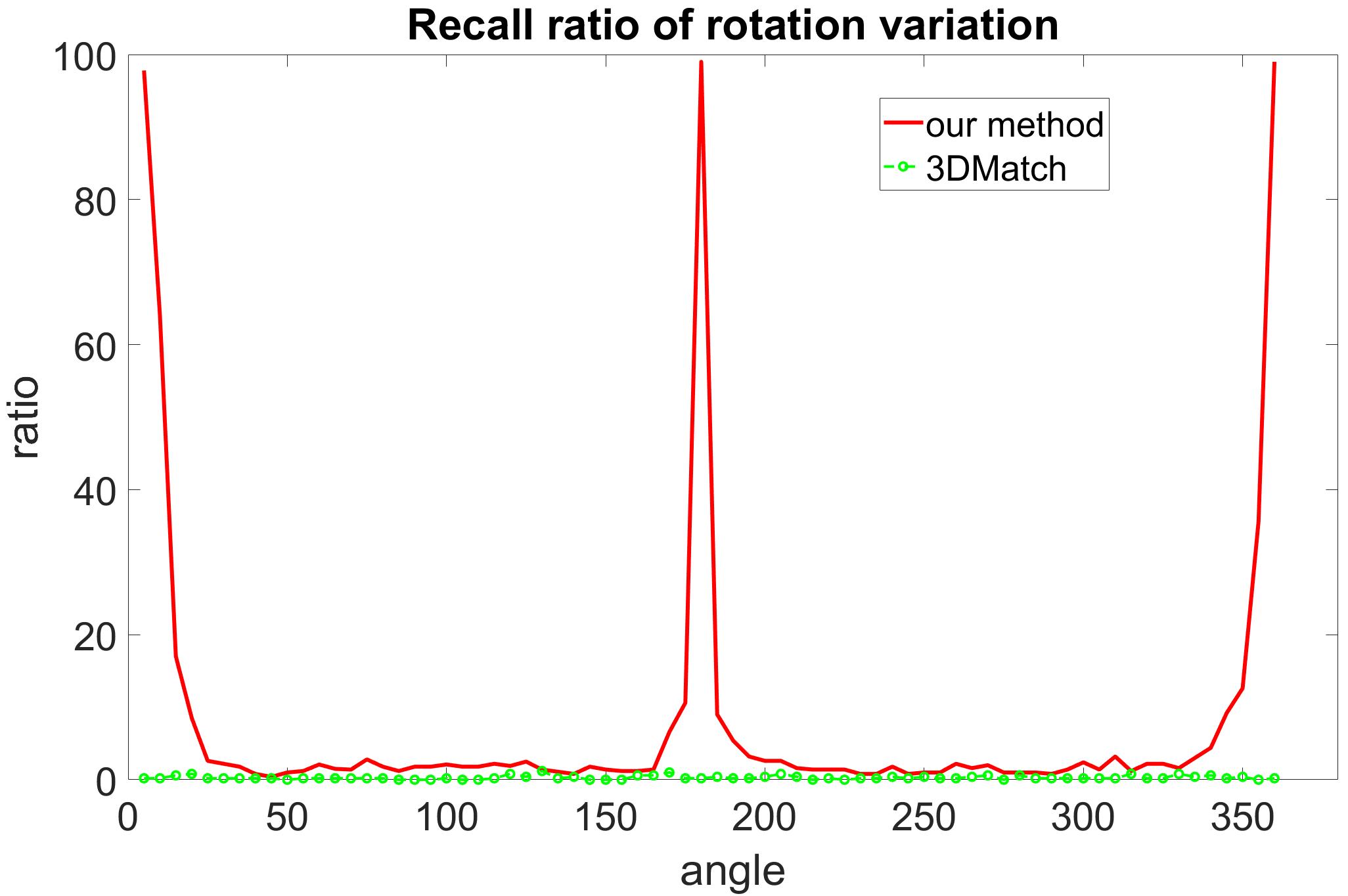}
	\caption{The recall ratio of ration variation.}
	\label{f5}
\end{figure}
\textbf{Rotation:} One point cloud (P1) is copied and rotated by every 5 degree to build another point cloud (P2). Totally, 72 sets of point cloud are produced. Each rotated point cloud with the original point cloud combined to build a pair with rotation variation. We sample 500 keypoints from P1 and record the index in the point cloud. In the rotated point cloud, I recall these keypoints by using these indexes. The groundtruth is the index correspondence.
I test the training detector on these 72 pairs. I use the recall ratio $R_1= R_0/R_N$ to evaluate the performance, where $R_0$ is the total number that the best matched descriptor are the groundtuth, $R_N$ is the point number. The results show in Figure \ref{f5}. It shows that the accuracy of our method shows more than 15\% within 10 degree rotation while 3DMatch only detect about 0.1\%. When the rotation angle goes higher, the recall ratio goes down to approximately 2\% while 3DMatch goes 0\%. The accuracy increases from 160$^\circ$ and reach the peak in 180 $^\circ$. The next 180$^\circ$ range goes similar cycle.

\begin{figure}[ht]
	\centering
	\includegraphics[width=8cm,height=6cm]{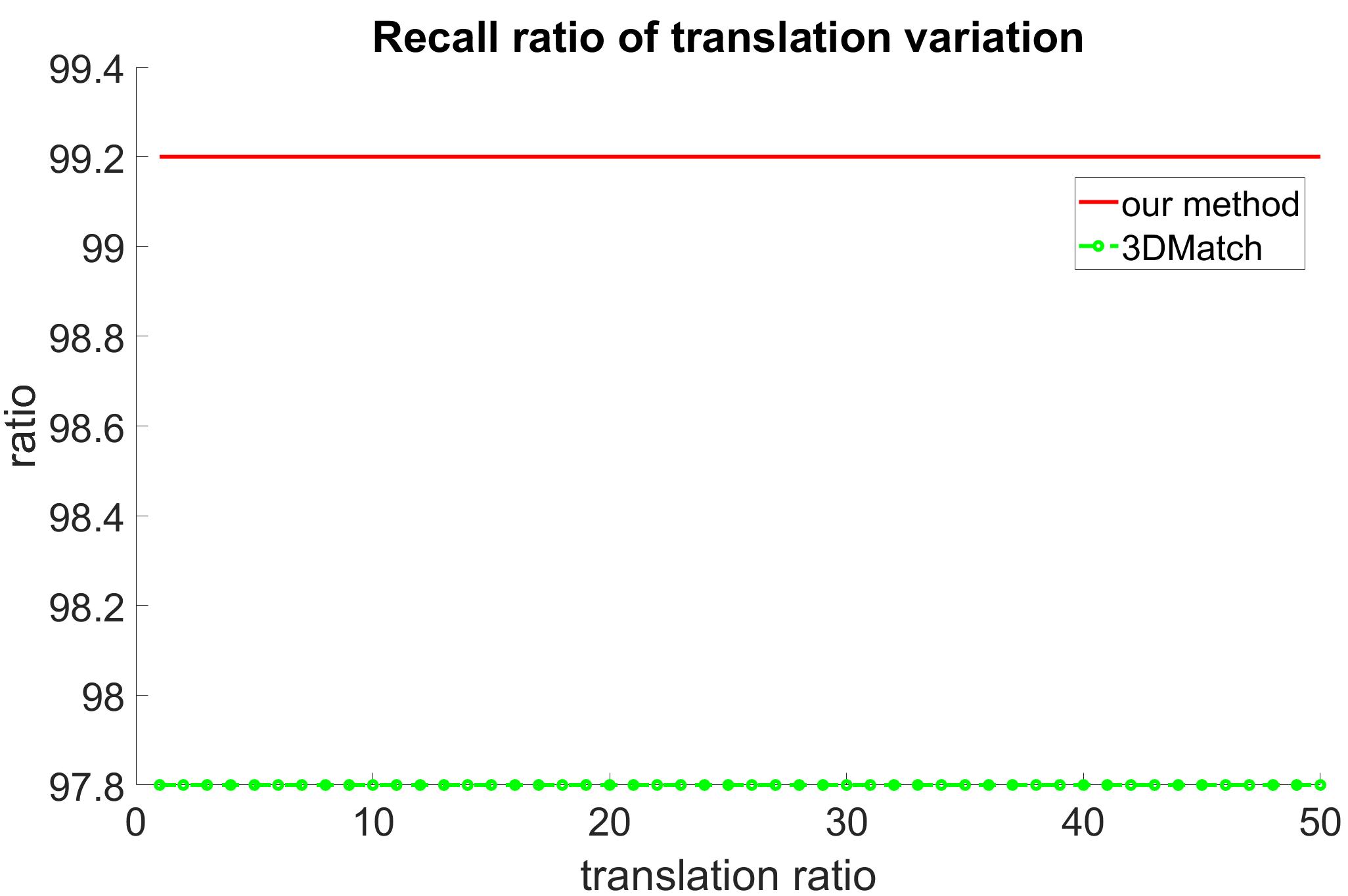}
	\caption{The recall ratio of translation variation.} 
	\label{f6}
\end{figure}

\textbf{Shifting on keypoints:} I shift keypoint randomly from 1\% to 50\% of the radius of point cloud containing sphere, totally 50 sets. If the point are shifted out of the containing sphere, we use the original point as the shifted point.   Figure \ref{f6} shows the results that our method shows high accuracy than 3DMatch.

\begin{figure}[ht]
	\centering
	\includegraphics[width=8cm,height=6cm]{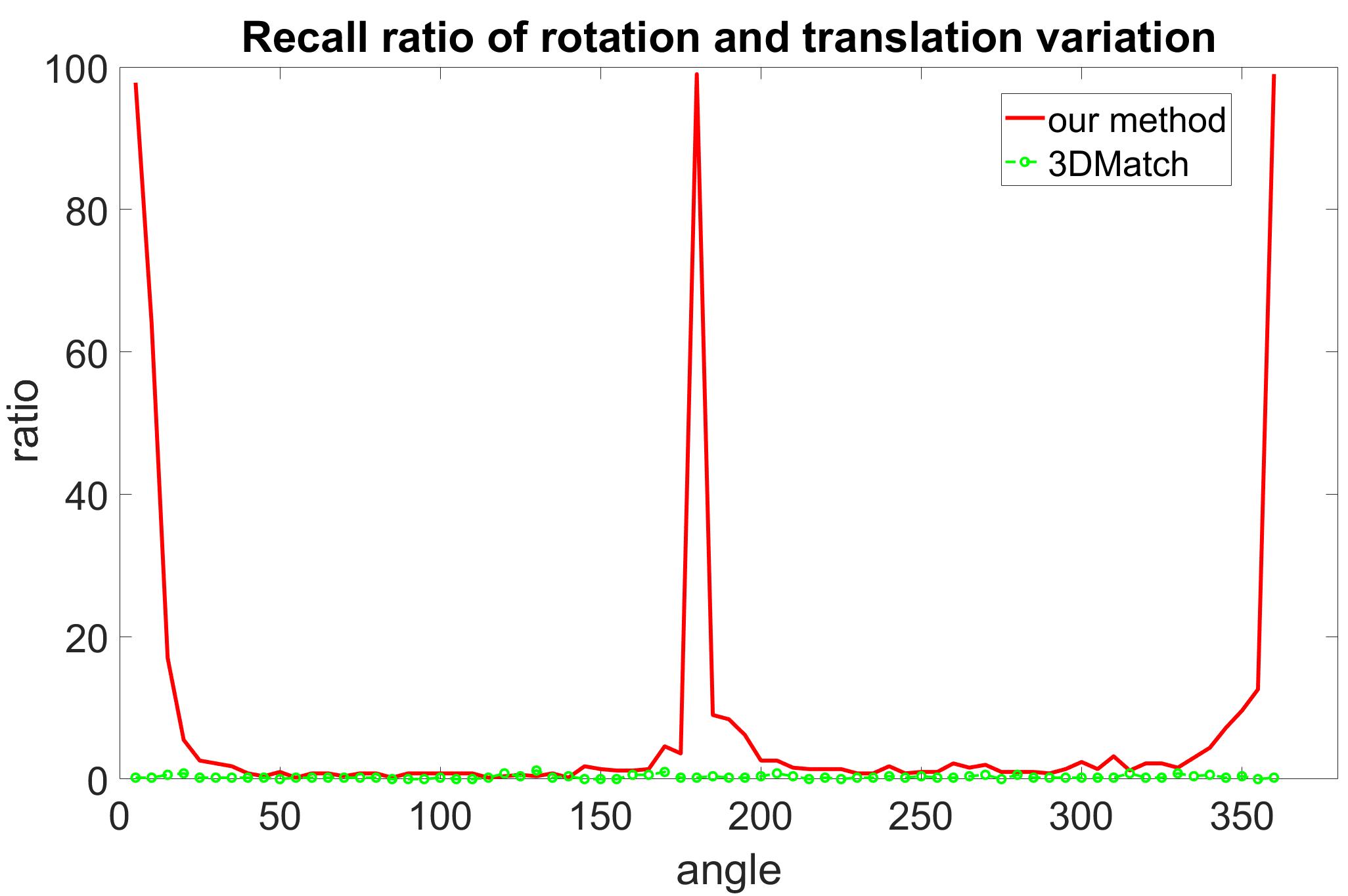}
	\caption{The recall ratio of rotation and translation variation.}
	\label{f7}
\end{figure}

\textbf{Rotation and Shifting:} I copy P1 to P2 and add each point in P2 a random shifting about 10\% of containing sphere. Then rotate P2 by every 5 $\circ$, totally 72 sets. Figure \ref{f7} shows our method obtains better results at most cases than 3DMatch. The proposed method achieves obviously better results within 20 degree rotation angle. Also, we obtain slightly better results than 3DMatch in angles larger than 20 degree rotation angle. It shows our ability in dealing with part of the rotation problem.

\section{Experiments on real 3D cross-source point cloud registration}
In this section, we will apply our network on real 3D cross-source point cloud registration and compare with 3DMatch \cite{3DMatch}. We capture one source of point cloud by using KinectFusion, and another source of point cloud by using struction-from-motion \cite{vsfm,dense}. These two sources of point clouds constitute cross-source point clouds. Following \cite{cs-coarse-to-fine}, we firstly normalize the scale variation. Because stable keypoints are very rare in cross-source point clouds, we use \cite{CSGM} to extract the structure points of the cross-source point clouds which is the stable information we currently know. With these structure points are extracted, we use the proposed method to extract the feature of these points and compare with 3DMatch.

\begin{figure*}[ht]
	\centering
	\includegraphics[width=18cm,height=7cm]{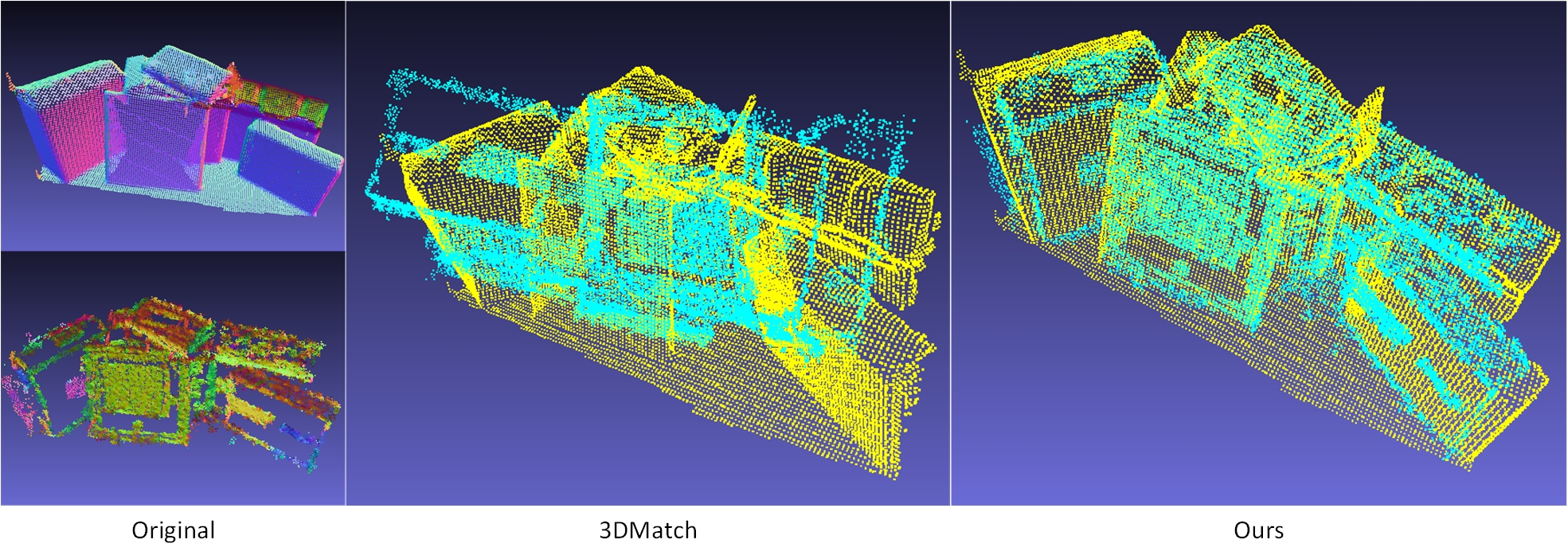}
	\caption{The registration result on cross-source 3D point clouds. It shows the original 3D cross-source point clouds, registration result of 3DMatch \cite{3DMatch}, registration result of the proposed method. }
	\label{f8}
\end{figure*}

Figure \ref{f8} shows the registration results of both methods. From Figure \ref{f8} we can see that the proposed method are much robust than 3DMatch in cross-source point cloud registration.

\section{Conclusion}
In this paper, we propose a method to learn a 3D descriptor for cross-source point cloud registration. We propose two strategies to train a robust neural network. Firstly, we use region-based samples to train the network. Secondly, we use data augmentation to extend our network be robust to rotation transformation variant. The experiments demonstrate that the proposed method output other method in terms of robustness and accuracy in the problem of cross-source point cloud registration.

\bibliographystyle{ieee}
\bibliography{egbib}

\end{document}